\documentclass[letterpaper, 10pt, conference]{ieeeconf}

\IEEEoverridecommandlockouts                              % This command is only needed if 
\usepackage{mydefs}		% Some of our own definitions

\begin{document}
\pagestyle{empty}

% TITLE
  \title{\LARGE \bf
  Night-to-Day Image Translation for Retrieval-based Localization
  }

  \author{
  Asha Anoosheh$^{1}$, Torsten Sattler$^{2}$,
                Radu Timofte$^{1}$, Marc Pollefeys$^{3,4}$,
                Luc Van Gool$^{1,5}$%
  \thanks{$^{1}$Computer  Vision  Lab,  Department of Information Technology \& Electrical Engineering, ETH Z\"urich, Switzerland}%
  \thanks{$^{2}$Dept. of Electrical Engineering, Chalmers University of Technology, Sweden}%
  \thanks{$^{3}$Department of Computer  Science, ETH Z\"urich, Switzerland}%
  \thanks{$^{4}$Microsoft, Switzerland}%
  \thanks{$^{5}$ESAT, KU Leuven, Belgium}%
  }%

%   \authorblockN{
%   \authorblockA{$^{1}$Computer  Vision  Lab,  Department of Information Technology \& Electrical Engineering, ETH Z\"urich}
%   \authorblockA{$^{2}$Computer  Vision  and  Geometry  Group,  Department of Computer  Science, ETH Z\"urich}
%   \authorblockA{$^{3}$Microsoft, Switzerland}
%   \authorblockA{$^{4}$ESAT, KU Leuven}
%   \authorblockN{{\tt\small ashaa@ethz.ch}, {\tt\small sattlert@inf.ethz.ch}, {\tt\small timofter@ethz.ch}, {\tt\small marc.pollefeys@inf.ethz.ch}, {\tt\small vangool@ethz.ch}}

%   Asha Anoosheh \\
%   Computer Vision Lab \\
%   D-ITET, ETH Z{\"u}rich \\
%   {\tt\small ashaa@ethz.ch}
%   \and
%   Torsten Sattler \\
%   Computer Vision Group\\
%   D-INFK, ETH Z{\"u}rich \\
%   {\tt\small sattlert@inf.ethz.ch}
%   \and 
%   Radu Timofte \\
%   Computer Vision Lab \\
%   D-ITET, ETH Z{\"u}rich \\
%   {\tt\small timofter@ethz.ch}
%   \and
%   Luc Van Gool \\
%   D-ITET, ETH Z{\"u}rich \\
%   ESAT, KU Leuven \\
%   {\tt\small vangool@ethz.ch}
%   \and
%   Marc Pollefeys \\
%   D-INFK, ETH Z{\"u}rich \\
%   Microsoft, Switzerland\\
%   {\tt\small marc.pollefeys@inf.ethz.ch}
  
  \maketitle

\begin{abstract}

Visual localization is a key step in many robotics pipelines, allowing the robot to (approximately) determine its position and orientation in the world.
An efficient and scalable approach to visual localization is to use image retrieval techniques. 
These approaches identify the image most similar to a query photo in a database of geo-tagged images and approximate the query's pose via the pose of the retrieved database image. 
However, image retrieval across drastically different illumination conditions, e.g. day and night, is still a problem with unsatisfactory results, even in this age of powerful neural models. This is due to a lack of a suitably diverse dataset with true correspondences to perform end-to-end learning. A recent class of neural models allows for realistic translation of images among visual domains with relatively little training data and, most importantly, without ground-truth pairings.

In this paper, we explore the task of accurately localizing images captured from two traversals of the same area in both day and night. We propose ToDayGAN -- a modified image-translation model to alter nighttime driving images to a more useful daytime representation. We then compare the daytime and translated night images to obtain a pose estimate for the night image using the known 6-DOF position of the closest day image. Our approach improves localization performance by over 250\% compared the current state-of-the-art, in the context of standard metrics in multiple categories.

\end{abstract}

%% ----------------------------------------------------------------------------
% BIWI SA/MA thesis template
%
% Created 09/29/2006 by Andreas Ess
% Extended 13/02/2009 by Jan Lesniak - jlesniak@vision.ee.ethz.ch
%% ----------------------------------------------------------------------------

\section{Introduction}
Many tasks such as autonomous vehicular navigation and mixed reality revolve around keeping track of the source of visual sensing, the camera, in its surroundings. The problem of being able to notice a previously observed spot is known as place recognition, and it is often intertwined with the related problem of localization: keeping track of one's position with respect to the previous position and the surrounding environment. Place recognition can aid or even serve as the basis for localization itself.

One way of performing place recognition is to directly compare traversal images against images captured during a potentially different traversal. Between comparisons, viewing conditions such as weather and lighting can change considerably. Ideally, places should be matched correctly regardless of differing conditions.
Yet in practice, existing methods are hampered by shifts between the domains of the images used for querying and those used for reference. Closing this domain gap should lead to an easier and more accurate comparison among images. This paper focuses on the problem of image comparison across contrasting visual conditions for the purpose of visual localization.

Modern learning-based methods such as deep neural networks should theoretically be well-suited for tackling the image comparison problem. But the issue holding them back currently appears to be a lack of appropriate training data. Training such a model for this task directly would require hundreds of thousands - or more - of images taken from many different positions, with multiple images taken at each position under diverse conditions, to ensure robustness to viewpoint and viewing condition changes. However, gathering or automatically generating this type of data is difficult.

Instead of gathering tedious quantities of labeled data, we exploit advantages of a recent class of neural networks that perform unpaired image-to-image translation~\cite{CYCLEGAN,NVIDIA,STARGAN,COMBOGAN,WESPE}. This refers to the idea of changing the visual properties of images from one domain to appear as if it came from another, where domains are defined by collections of data alone. The process is conveniently unsupervised: rather than requiring tuples of images depicting the same place under different conditions, one just needs collections of any images taken under the same conditions. Based on these collections, one can train a model that translates between the different conditions. In addition, these models can, in our experience and purposes, produce visually-adequate results with as little as $\sim$500 points of data per domain, unlike most neural network-based tasks which demand tens-of-thousands to millions.

Image translation began as altering the characteristics of an image between perceived styles for artistic and/or entertainment purposes. With projects such as~\cite{CYCLEGAN,STARGAN,NVIDIA,COMBOGAN,SMIT} breaking ground, it was now possible to perform high-quality image translation.
Soon thereafter, the idea was used to aid other learning tasks \cite{NVIDIA,CYCADA,GTA,GTA2,OXFORD}. These works show that being able to attain high-quality representations of images in the appearance of other domains is useful for tasks containing a shift in the data domain; irrelevant source-domain-specific information is discarded and helpful target domain-specific details can be filled in.

This paper explores different methods for tackling the problem of similarity between images captured from car-mounted cameras, specifically for applications in autonomous driving. Our method involves performing image translation from a source (night) to a target domain (day) and feeding the output to an existing image comparison tool. Using a fixed representation for comparing the images allows us to decouple the problem into domain adaptation and image matching, where we only need to focus on the former. The number of parameters and degrees of freedom is thus reduced. Additionally, the objective of the translation model is modified so that the output not only contains the visually perceptible qualities of the target domain but also the properties that the image comparison tool relies on most heavily. We present an image-translation model that specializes discriminators to directly address a task. It is also one of the first instances of applying image translation to the problem of retrieval-based localization. Our approach greatly outperforms all current state-of-the-art methods on a challenging benchmark.

\section{Related Work}

\subsection{Image-to-Image Translation}
Many tasks in computer vision can be thought of as translation problems where an input image $a$ is to be translated from domain $A$ to $b$ in another domain $B$. Instead of sampling from a probability distribution to generate images as with regular Generative Adversarial Networks (GANs)~\cite{GAN}, translation approaches produce an output conditioned on a given input~\cite{PIX2PIX,PIX2PIXHD}.

\begin{figure}
\begin{center}
\includegraphics[width=\linewidth]{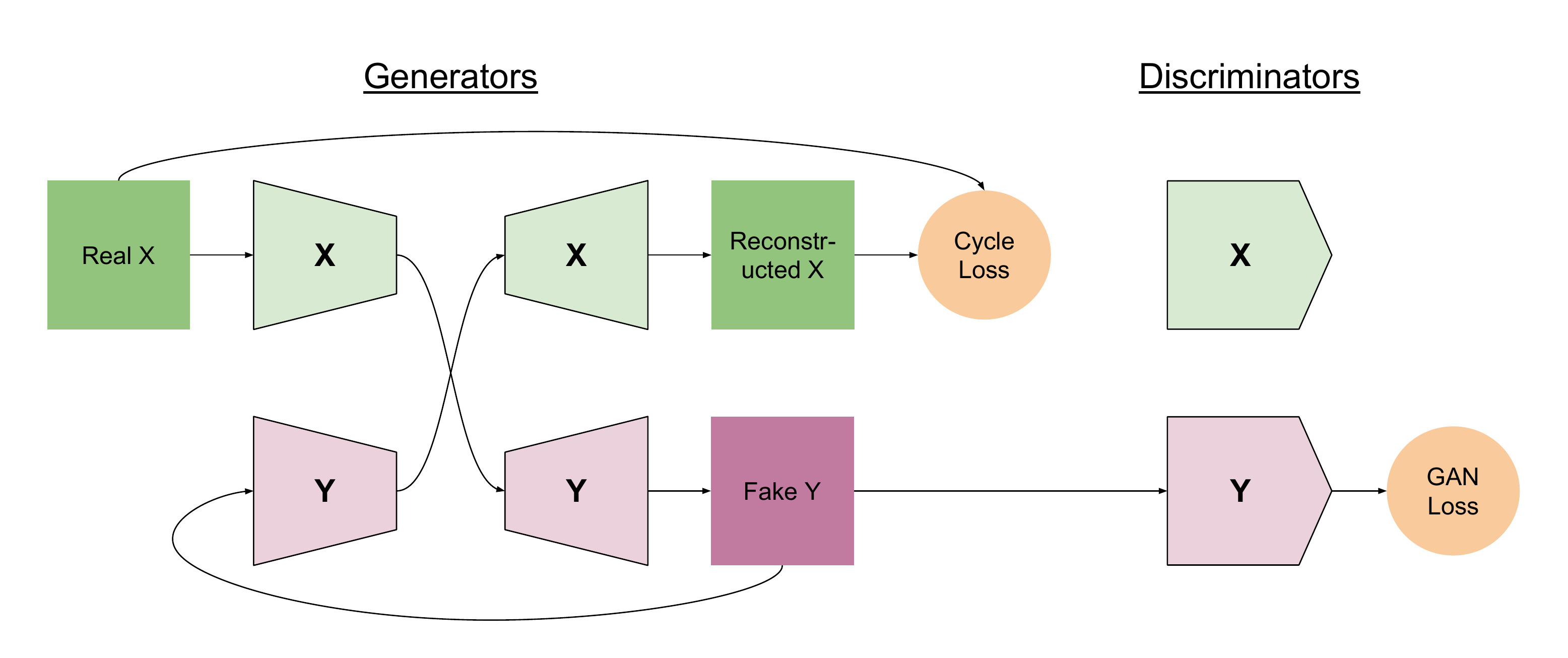}
\end{center}
   \caption{CycleGAN/ComboGAN generator training pass for domain direction $A \rightarrow B$. This pass is repeated symmetrically for direction $B \rightarrow A$ as well.}
   \label{fig:training}
\end{figure}

Introduced by Zhu~\etal, CycleGAN~\cite{CYCLEGAN} extends image-to-image translation to an unsupervised framework, meaning no alignment of image pairs is necessary. It relies on GANs, a class of neural networks proven to be excellent for capturing the rich distributions of natural images~\cite{GAN}. Adversarial training involves a generator and discriminator network, updated in alternating steps, allowing both to gradually improve alongside each other; first D is trained to distinguish between one or more pairs of real and generated samples, and then the generator is trained to fool D with generated samples.
CycleGAN consists of two pairs of generator and discriminator nets, $(G_A, D_A)$ and $(G_B, D_B)$, where the translators between domains $A$ and $B$ are $G_A : A \rightarrow B$ and $G_B : B \rightarrow A$. $D_A$ is trained to discriminate between real images $a$ and translated images $G_B(b)$, while $D_B$ is trained to discriminate between images $b$ and $G_A(a)$. The system is trained using both an adversarial loss and a cycle consistency loss (see Figure \ref{fig:training}). The Cycle consistency loss is a way to regularize the highly unconstrained problem of translating an image unidirectionally, by encouraging mappings $G_A$ and $G_B$ to be inverses such that $G_B(G_A(a)) \approx a$ and $G_A(G_B(b)) \approx b$. The full CycleGAN objective is expressed:

\small
\begin{multline} \label{eq:lsgan_loss}
\mathcal{L}_{GAN}(G,D,a,b) = 
	\mathbb{E}_b [(D(b) - 1)^2] + \mathbb{E}_a [D(G(a))^2]
\end{multline}
\vspace{-14pt}
\begin{multline} \label{eq:cycle_loss}
\mathcal{L}_{cyc} =
	\mathbb{E}_a [||G_B(G_A(a)) - a||_1] + \mathbb{E}_b [||G_A(G_B(b)) - b||_1]
\end{multline}
\vspace{-14pt}
\begin{multline} \label{eq:cyclegan_loss}
\mathcal{L}_{CycleGAN} = \mathcal{L}_{GAN}(G_A,D_B,a,b) \\
+ \mathcal{L}_{GAN}(G_B,D_A,b,a) + \lambda\,\mathcal{L}_{cyc}
\enspace.
\end{multline}
\normalsize

Subsequent works improved upon this foundation. Liu~\etal used a variational autoencoder loss formulation to encourage a shared feature space for a model they named UNIT~\cite{NVIDIA}. Ignatov~\etal performed one-way translation for image enhancement, using two discriminators per domain - one for color and the other for texture - rather than just one~\cite{WESPE}. And ComboGAN~\cite{COMBOGAN} and SMIT~\cite{SMIT} allowed for $n$-domain translation, solving the exponential scaling problem in the number of domains.

This project uses ComboGAN as the base image-to-image translation model, which is equivalent to CycleGAN in the case of two domains only. And since we do not work with more than two domains in this paper, it's irrelevant which one is used as the starting point (though using ComboGAN means the model can automatically serve more then two domains if need be). We modify ComboGAN incrementally over the course of our experiments to better suit our exact task. We also compare results with UNIT to see how another translation model performs, out-of-the-box.

\subsection{Place Recognition and Localization} \label{placerecog}
Place recognition refers to the task of identifying a real world location from images of said place - essentially location classification. Ideally this process should be invariant to various image and world properties such as camera position, orientation, weather conditions, etc. Visual localization is the process of identifying the camera location (and sometimes orientation) either relative to a local or global map. One way of achieving this is via image retrieval: finding the most similar image with a known pose to a unknown query. In such a case, invariance to camera pose is not desired, as these are critical to calculating pose as accurately as possible.

A traditional tool widely used in place recognition and/or image retrieval is the VLAD descriptor~\cite{VLAD}. The descriptor is a rather low-dimensional vector intended to serve as a featurization of the image as a whole. A visual vocabulary is built from a diverse dataset, extracting $D$-dimensional descriptors from affine-invariant detections - and clustered into $k$ centers. Afterwards, for a given query image with $n$ local descriptors, the residual from each descriptor to each cluster center is calculated and summed per cluster, resulting in $k$ $D$-dimensional aggregate vectors. These are then concatenated and normalized to form a unit-length, $k$$\times$$D$-dimensional VLAD descriptor. Vectors across images can now be compared directly (using the original Euclidean distance metric involved) to quantify similarity.

DenseVLAD, a modification of the VLAD procedure, was formalized by Torii~\etal~\cite{DENSEVLAD} to improve place recognition as a subprocedure for camera pose estimation. Instead of using a detector to choose where to extract features, 128-dimensional RootSIFT descriptors \cite{ROOTSIFT} - an adjustment to the classic SIFT \cite{SIFT} - are simply extracted on a regular grid throughout the image. This removes the issue of inconsistent detections affecting the process altogether. The next steps follow the standard VLAD clustering with $k$ clusters to produce a $128k$-dimensional VLAD descriptor per image, which is then down-projected via PCA.
% SIFT descriptors are a modified binning of gradients surrounding a point of interest. First a $N \times N$ grid (typically $N=16$) of x-\&-y gradients $I_x, I_y$ are computed around a pixel, followed by their magnitude and orientation ($\sqrt{I_x^2 + I_y^2}$ and $atan2(I_y, I_x)$, respectively). Each smaller $n \times n$ block (typically $n=\sqrt{N}$) within the grid is grouped into a histogram of orientations weighted by magnitude. After clipping and normalization, we get $N$ $n \times n$ spatial histograms to describe the image region. DenseVLAD extracts multiple versions of these descriptors per image using different bin sizes $n$ for the SIFT histograms, in order to describe the image at multiple feature scales.

In 2015, NetVLAD~\cite{NETVLAD} reformulated the VLAD process into a neural network framework. The idea was to maintain the clustering and unit vector principles while allowing for a differentiable, trainable architecture. It consists of a feature-extraction portion, which could be early layers from a pre-trained convolutional network, and a ``VLAD-Core'' module which performs soft-clustering to output a VLAD-like vector.
% 
% In this paper, we apply DenseVLAD directly in our approach to produce representations of images that can be directly compared using a metric. It matches images after we process them in various ways, as DenseVLAD (assuming an existing vocabulary) requires no training and can be applied to any single-channel input.

\subsection{Image Translation for Visual Localization} \label{rel_oxford}
Domain shift, or equivalently, dataset bias, is an often encountered problem in methods that learn from data. This occurs when a model built from data is applied to data that differs in some characteristics. Differences between training and inference data can cause large variation in outputs if models are not robust against them. Domain Adaptation is the practice of mitigating the effects of domain shifts~\cite{ADDA,DANN}. Image-based Localization under differing visual conditions has been shown to be heavily affected by domain shifts~\cite{TORSTEN}.

A concurrent work, published during the development of this paper, by Porav~\etal~\cite{OXFORD} sought to use CycleGAN to approach almost the same problem as ours: effectively comparing images under different illumination conditions by visually translating one domain to the other  - for example changing nighttime images to day prior to feature matching. They add additional cycle losses on top of the original CycleGAN setup: constraints that reconstructed images should have identical second-order derivatives and Haar responses. This is because the localization procedure employed utilizes SURF~\cite{SURF} features, which rely on second order-derivatives for detection and Haar responses for description. The model was tested on localizing a night sequence to a daytime sequence from the Oxford RobotCar dataset \cite{ROBOTS}, synthesizing daytime images from the night ones to match features against the real daytime images.
This is closely related to our main approach, where we translate images to improve descriptor matching. Whereas their method enforces a descriptor-aiding feature loss on the input and cyclically-reconstructed image in hopes of these features staying present in the intermediate translated image, our main approach in this paper enforces a similar type of loss directly on the initial translated output. Their method also used a subset of data from the same dataset as ours, albeit trained on a different set of cameras with different orientations and intrinsics. Additionally, it was evaluated on the related - but not directly comparable - task of visual odometry on sequential data with synchronized starting positions. We attempted to adapt the process to our task, but had to conclude that a fair comparison was not possible in the scope of this work. %can be made due to various factors. %The process could have been adapted for our data, if not for lack of source code.

\section{Our Method - ToDayGAN} \label{method}

\begin{figure}[t!]
\begin{center}
\includegraphics[width=0.88\linewidth]{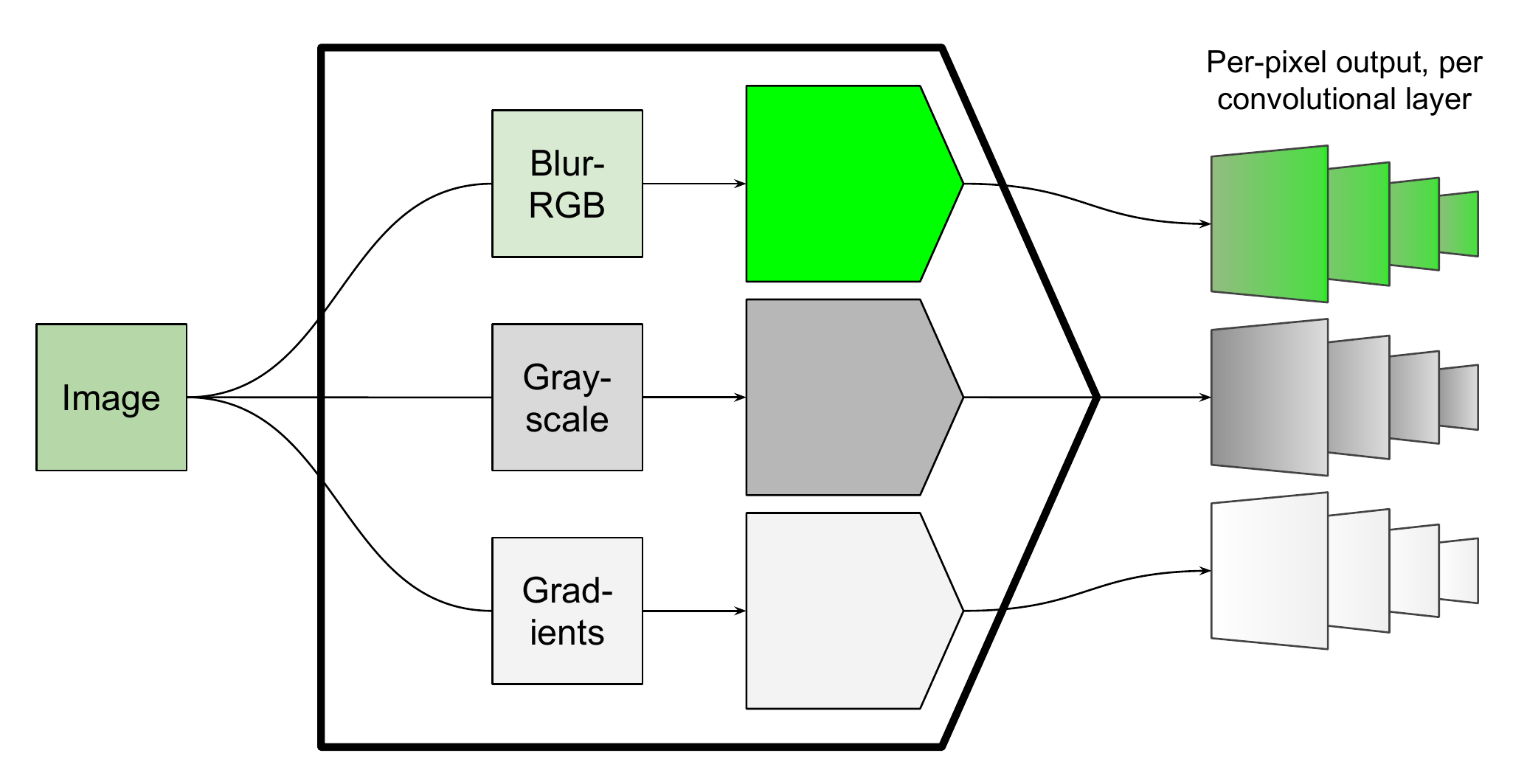}
\end{center}
   \caption{Each discriminator in our ToDayGAN is comprised of three network clones that operate separately on the blurred-RGB, Grayscale, and $xy$-Gradients of its input image. Each discriminator outputs a decision at each convolutional layer to cover different receptive-field sizes.}
   \label{fig:discriminator}
\end{figure}

Our approach, ToDayGAN, to tackle day-night localization between a daytime reference set with known poses and a nighttime query set is as follows. First, an image translation model is trained to translate between day and night image domains. Second, the night-to-day direction is used to transform nighttime images into a daytime representation. Both the translated images and the reference images are then fed to an existing featurization process to obtain a feature vector per image. Nearest neighbor search gives us a closest matching reference image per query. The pose of the query image is then  approximated by the pose of the nearest neighbor daytime image.

Our image featurization tool used is DenseVLAD~\cite{DENSEVLAD}. As detailed in Section~\ref{placerecog}, DenseVLAD is an improved version of the VLAD image description technique for describing and comparing image data. By densely extracting descriptors from images and forgoing the detection stage, DenseVLAD is more robust to strong appearance changes than standard VLAD is. As shown in~\cite{TORSTEN}, DenseVLAD still outperforms more modern methods such as NetVLAD in terms of generalization on day-night image matching.

Our image-translation model is built using the image-translation model ComboGAN~\cite{COMBOGAN} as its base. The generator networks are identical to the networks used in CycleGAN (see Table~A.1 in \cite{COMBOGAN} for details on the network architecture), yet each is divided in half, the frontal halves being encoders and the latter halves decoders. For the case of two domains, ComboGAN's structure and training procedure are identical to CycleGAN's; thus it is irrelevant which one is used as the starting point, though using ComboGAN means the model can automatically serve more than two domains if need be.
Instead of just using ComboGAN as is, we adjust its setup to fit our problem. More specifically, ComboGAN's discriminators are modified, resulting in a noticeable improvement on the localization task performed using the translated images.

Following WESPE~\cite{WESPE}, we replicate the discriminators to specialize on different aspects of the input. In the complete version of ToDayGAN, each domain's discriminator (see Figure~\ref{fig:discriminator}) contains \emph{three} copies of the network, expanding on insights from WESPE. One takes the luminance (grayscale) of the input image, one takes the RGB image blurred by a 5x5 3$\sigma$ Gaussian kernel (exactly as in~\cite{WESPE}), and the last takes the horizontal/vertical gradients of the image. These three discriminators can now separately focus on texture, color, and gradients. Each one is equal in architecture and hyperparameters, and their losses are averaged in the end equally. As shown by our experiments, with the addition of each discriminator comes a significant performance boost.

The third discriminator is a novel contribution that serves to emulate the process of extracting SIFT descriptors. The DenseVLAD implementation from~\cite{DENSEVLAD} converts the input image to grayscale and  downsamples it by a factor of $2\times$ by skipping every other pixel. Then it creates histograms of magnitude-weighted gradient orientations after computing gradients via convolution with a $[-1\ 0\ 1]$ kernel for $x$-direction gradients and its transpose for the $y$ direction. Therefore our model uses a $1\times1$ stride-2 convolution to obtain the downsampled image and convolves it with the two filters to obtain the same $xy$ gradients for the discriminator in a differentiable manner. As opposed to~\cite{OXFORD}, we use this discriminator to create matching-relevant features in the translated version that were nonexistent in the original, whereas they simply preserve the relevant features from the original images in their cyclic reconstructions.

We also attempted to  emulate the process of  computing DenseVLAD descriptors in more detail by including gradient magnitudes and orientations in the  discriminators.
However, experimental results showed a worse performance (by about a factor of $2\times$) compared to using gradients only.

In addition, the discriminators output a label/decision after each downsampling layer. Being able to discriminate images at multiple scales encourages consistency in both low- and high-level image statistics, rather than just at the final arbitrary receptive field size. This idea was inspired by \cite{PIX2PIXHD}, where multiple discriminators see differing sizes of the input image; instead, in our case, single discriminators output multiple decisions. The outputs for the final loss are weighed linearly, ascending toward the last layer, as the complexity and power of the network's predictions increase with depth. This can be seen as the $n$ outputs, in ascending layer order, weighed by $[1,2,..,n]$ then summed and divided by $n \sum_i^n i$ to average out.

A recent discriminator loss format, the Relativistic Loss, was introduced~\cite{RELATIVE}, which alters the discriminator loss formulation by only requiring it to determine whether an input is more real relative to a fake, rather than to determine realness in an absolute manner. The motivation behind this is to stabilize training overall by preventing the discriminator to become too powerful in relation to the generator. Equation \eqref{eq:rel_loss} defines ComboGAN's least-squares GAN loss from Equation \eqref{eq:lsgan_loss} newly adapted to the Relativistic Loss formulation: 

\begin{equation} \label{eq:rel_loss}
\resizebox{0.94\hsize}{!}{$
\mathcal{L}_{GAN}(G_A,D_B,A,B) = 
        \begin{cases}
            \mathbb{E}_a\mathbb{E}_b [ (D_B(b) - D_B(G_A(a)) - 1)^2 ] \\\qquad \text{\textit{from Discriminator perspective}} \\
            \mathbb{E}_a\mathbb{E}_b [ (D_B(G_A(a)) - D_B(b) - 1)^2 ] \\\qquad \text{\textit{from Generator perspective}}
        \end{cases}
$}
\end{equation}

\section{Experimental Setup}
\subsection{Dataset}
Our source of images for training and evaluation is the Oxford RobotCar dataset~\cite{ROBOTS}. It contains multiple video sequences of the same 10km route captured from an autonomous vehicle in Oxford, England. Three Point Grey Grasshopper2 cameras were mounted on the left, right, and rear of the vehicle, and the traversals were taken over the course of a year, providing variation in lighting, time-of-day, and weather. Though it resulted in over 20 million 1024$\times$1024-resolution images in total, only a subset of certain traversals are used for our experiments. The image sets contain corresponding left, right, and rear views, meaning there exist an  image triplet per timestamp.

We use the RobotCar Seasons variant~\cite{TORSTEN} for evaluation, which provides accurate camera poses for a set of reference and query images:  
A subset of 6,954 camera triplets of the original RobotCar dataset, which we refer to as Day (known as ``overcast-reference'' in~\cite{TORSTEN}), is used as a reference. Another set of 438 triplets, which are captured at night are used as a query set (``night'' in~\cite{TORSTEN}). We also use a second query set of 440 images captured at night during rain (``night-rain'' in~\cite{TORSTEN}), whose only purpose is to examine the transferability of our technique (trained without any rain) to a visually-different domain. Finally, another set of 6,666 nighttime triplets, not included in~\cite{TORSTEN} and used only for training, is randomly sampled from three other traversals of the RobotCar dataset directly. We call these three nighttime datasets Night-query, Night-rain, and Night-train.

The same Day image set is used during training and testing of ToDayGAN. 
This is intentional as inference is only ever performed on Night images and since localization is an instance-level task, \ie, reference images representing the scene are always available. The goal of the model is to generalize on Night inputs and specialize the outputs to the specific Day reference dataset from which the poses were precalculated. Hence, only the Night images are separated into training and testing. We make the assumption that, due to the relative ease of gathering unlabeled data for ToDayGAN, others making use of this procedure can easily tailor the training process for the specific visual qualities of their own reference dataset.

Our datasets are subsets of much larger original video sequences. In~\cite{TORSTEN}, they have been sub-sampled to reduce sizes to manageable levels. A 3D map was constructed from a vehicle's traversal so that the pose of each image can be estimated~\cite{TORSTEN}, and then an image was sampled every meter. For the daytime images, using the vehicle's inertial navigation system and 3D visual tracking is sufficient to build a map, but LIDAR data (also captured by the same vehicle) was necessary to obtain ground-truth poses for nighttime images. We use  poses provided by~\cite{TORSTEN}.

As side-view images are only used to boost data count during some of the training runs, side-views for the query-Nighttime remain unused, and the correspondence (or lack thereof) for the three views is irrelevant for the purpose of our experiments. As the problem is formulated below, the same Daytime images are available during both training and inference, so the same images are used for both, meanwhile the night images are independent in all stages. Exact details of each dataset can be found in Table~\ref{tbl:robotcar_details}.

\begin{table}
\centering
\caption{Details of RobotCar and RobotCar Seasons dataset used.}
		\label{tbl:robotcar_details}
\setlength{\tabcolsep}{4pt}
\begin{tabular}{|l|c|c|c|}
\hline
Condition & Purpose & Recorded & \# triplets \\
\hline\hline
overcast & reference \& training & 28 Nov 2014 & 6,954 \\
night & training & 27 Feb \& 01 Sep 2015  & 6,666 \\
\hline
night & query & 10 Dec 2014 & 438 \\
night-rain & query & 17 Dec 2014 & 440 \\
\hline
\end{tabular}
\end{table}

\subsection{Training Setup}
The following apply to all three types of our models. Images, unless mentioned that left and right viewpoint images were used from the RobotCar dataset, were trained on rear views only. And unless stated otherwise, images in our trials were scaled to $286\times286$ size and randomly cropped to $256\times256$ for training. If a $512\times 512$ resolution is used, training crops are of size $384\times384$ due to memory constraints. Memory also restricts training on resolutions higher than $512\times 512$. Inference is always on the pre-crop size because our fully-convolutional architecture allows for arbitrary input sizes. Batches are not used, and random image flipping (left-right) is enabled. Training is run for 40 epochs. Learning rates begin at $2\text{e-}4$ for generators and $1\text{e-}4$ for discriminators, are constant for the first half of training and decreasing linearly to zero during the second half. The $\lambda$ from equation~\eqref{eq:cyclegan_loss} is set to 10.0, as in \cite{CYCLEGAN}.

DenseVLAD uses the $k=128$ pretrained cluster centers provided by~\cite{DENSEVLAD}. The VLAD vectors are projected down to 4096 dimensions via PCA prior to comparisons. We also keep the default SIFT extraction scales used in DenseVLAD, at $n \in \{4, 6, 8, 10\}$.

\begin{figure}
\centering
\setlength{\tabcolsep}{1pt}
\renewcommand{\arraystretch}{1}
\resizebox{\linewidth}{!}
{
\begin{tabular}{cccc}
\rotatebox{90}{\parbox{2cm}{real nighttime}} &
\includegraphics[width=.4\linewidth]{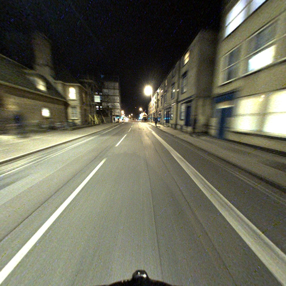}&
\includegraphics[width=.4\linewidth]{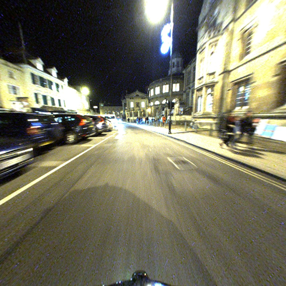}&
\includegraphics[width=.4\linewidth]{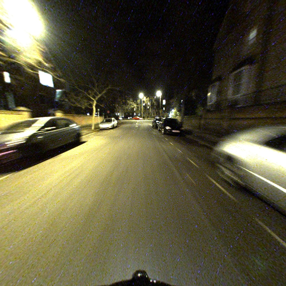}
\\
\rotatebox{90}{\parbox{2cm}{ToDayGAN (ours)}} &
\includegraphics[width=.4\linewidth]{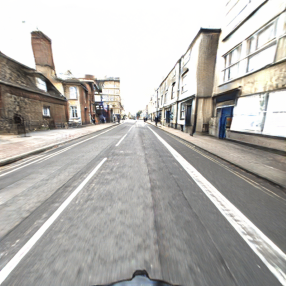}&
\includegraphics[width=.4\linewidth]{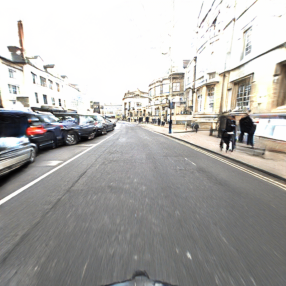}&
\includegraphics[width=.4\linewidth]{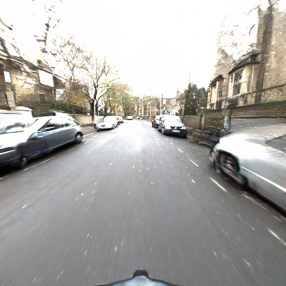}\\
\rotatebox{90}{\parbox{2cm}{ToDayGAN w/o Rel-Loss}} &
\includegraphics[width=.4\linewidth]{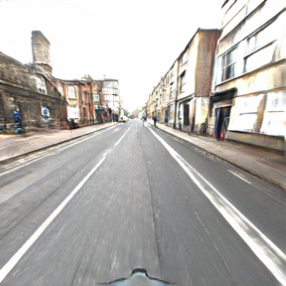}&
\includegraphics[width=.4\linewidth]{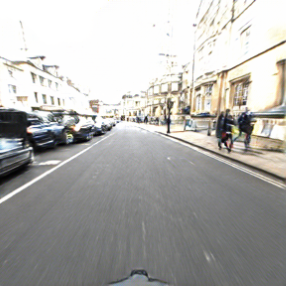}&
\includegraphics[width=.4\linewidth]{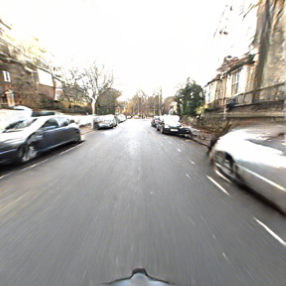}\\
\rotatebox{90}{\parbox{2cm}{UNIT~\cite{NVIDIA}}}&
\includegraphics[width=.4\linewidth]{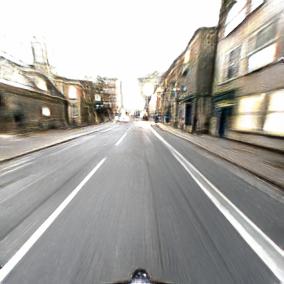}&
\includegraphics[width=.4\linewidth]{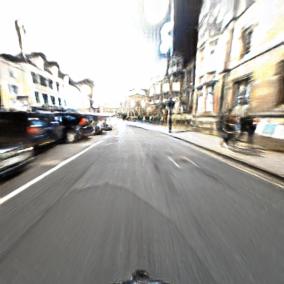}&
\includegraphics[width=.4\linewidth]{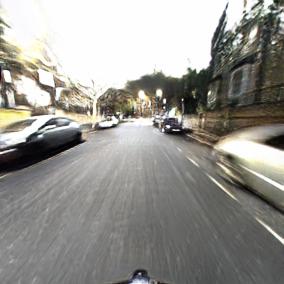}
\end{tabular}
}

\caption{Top to bottom row: Example real night images before translation, synthetic day images using the best-performing model, same model but without the Relativistic Loss, and synthetic day images produced by UNIT.}
\label{fig:res_real}
\end{figure}

\subsection{Evaluation Protocol \& Baselines}
Following the evaluation protocol of~\cite{TORSTEN}, we report the percentage of query images whose predicted 6-DOF poses is within three error tolerance thresholds: 5-meter and 10-degree, 0.5-meter and 5-degree, and 0.25-meter and 2-degree. Evaluating on the RobotCar Seasons dataset proposed in~\cite{TORSTEN} enables us to directly compare our approach to state-of-the-art methods.

Table~\ref{tbl:base_results} lists the results obtained by running DenseVLAD matching on the Daytime and Nighttime images (including Night-rain) directly to benchmark what we believe according to~\cite{TORSTEN} to be the best known solution until now. We include additional results from histogram-equalization of images prior to matching; we try this both on the query images only and also on both query and reference.
We also compare with out-of-the-box CycleGAN and UNIT~\cite{NVIDIA}. Note that UNIT is not tailored to any task other than perceptive quality of images translated.

Lastly, we compare the best results obtained using our methods with the state-of-the-art methods found in~\cite{TORSTEN}. These include structure-based localization techniques ActiveSearch~\cite{ACTIVESEARCH} and CSL~\cite{CSL}, in addition to the image-based FAB-MAP~\cite{FABMAP} and NetVLAD~\cite{NETVLAD}. Note that \cite{TORSTEN} report that they were not able to train pose regression techniques such as PoseNet \cite{POSENET} on the dataset. On the night query set, structure based methods performed very poorly; meanwhile, image-based approaches attained the highest accuracies for the same criteria. DenseVLAD was the best of all methods, followed by NetVLAD.

\begin{table}
\centering
\caption{DenseVLAD night query results for direct raw-image baselines, images histogram-equalized beforehand, and those translated with UNIT.}
		\label{tbl:base_results}
\setlength{\tabcolsep}{4pt}
\begin{tabular}{|c|c!{\vrule width 2pt}c|c|c|}
\hline
 &  &   \multicolumn{3}{c|}{Threshold Accuracy (\%)} \\
Preprocess & Resolution & 5m \ang{10} & 0.5m \ang{5} & 0.25m \ang{2} \\
\hline\hline
Direct & 286 & 13.9 & 2.9 & 0.4 \\
Direct & 512 & 10.2 & 1.8 & 0.2 \\
Direct & 1024 & 19.9 & \textbf{3.4} & \textbf{0.9} \\
\hline
Hist-Eq night only & 1024 & \textbf{23.7} & 2.5 & 0.7 \\
Hist-Eq night \& day & 1024 & 16.7 & 2.7 & 0.4 \\
\hline
CycleGAN~\cite{CYCLEGAN} & 286 & 10.7 & 0.9 & 0.0 \\
UNIT~\cite{NVIDIA} & 286 & 17.1 & 0.4 & 0.0 \\
\hline
\end{tabular}
\end{table}

\section{Results} \label{results}
Table~\ref{tbl:main_results} shows the localization rates gathered over various configurations of our own model to determine the effect of certain modifications and variables. Sample visuals from the process can be seen in Figure \ref{fig:res_real}.
These include a version of the model with only one discriminator per domain but with all three input features (color, texture, and gradients) concatenated along the channel dimension as a single larger input. This can signal whether separate models are needed, or if simply having these features pre-extracted is the key.

The ``Discriminators'' column contains up to three letters representing the types of discriminators used for that trial. ``C'' stands for Color, ``L'' for Luminance, and ``G'' for Gradients. Note that when ``C'' is not used in conjunction with an ``L'', the RGB image remains un-blurred when input to the color-discriminator. Likewise, when ``L'' is not used in conjunction with a ``C'', the entire model is run in grayscale, for obvious reasons. ``C+L+G'' represents unification as a single discriminator with the inputs concatenated. And lastly a trial was performed where the gradient discriminator received as input the magnitude and orientation of gradients in place of the gradients themselves, denoted by ``M''. This is closer to the actual DenseVLAD process and should \emph{theoretically} be better suited as task adaptation.

``Rel.-Loss'' in Table~\ref{tbl:main_results} means the discriminator loss is the Relativistic Loss mentioned in Section~\ref{method}. ``L/R'' indicates whether the left and right camera images were used to enlarge the training set.
``Dual-Eval.'' refers to our Dual-Evaluation procedure added as a finishing-touch enhancement to our models. For this, a horizontally-flipped version of each query image is fed to the network and then re-flipped for DenseVLAD featurization. Distances are calculated between these and the references as well, and the nearest neighbor is the minimum of these and the original unflipped distances. As the network is not invariant to left-right mirroring, this produces two similar yet different ``opinions,'' which boosts accuracies in our case.

\begin{table}
\centering
\caption{Ablation study for the proposed method.}
		\label{tbl:main_results}
\setlength{\tabcolsep}{4pt}
\resizebox{1.02\columnwidth}{!}{
\begin{tabular}{|c|c|c|c|c!{\vrule width 2pt}c|c|c|}
\hline
 &  &  &  &  &  \multicolumn{3}{c|}{Threshold Accuracy (\%)} \\
Resolution & Discriminators & Rel.-loss? & Dual-Eval.? & L/R? & 5m \ang{10} & 0.5m \ang{5} & 0.25m \ang{2} \\
\hline\hline
286 & C  &  &  &  & 12.1 & 1.8 & 0.2 \\
286 & C, G  &  &  &  & 28.9 & 6.4 & 0.9 \\
286 & L, G  &  &  &  & 9.1 & 2.3 & 0.4 \\
286 & C, L  &  &  &  & 16.2 & 2.5 & 0.6 \\
286 & C+L+G  &  &  &  & 18.7 & 2.5 & 0.4 \\
286 & C, L, M  &  &  &  & 15.0 & 1.8 & 0.4 \\
286 & C, L, G  &  &  &  & 30.8 & 5.9 & 0.6 \\
286 & C, L, G  & \checkmark &  &  & 36.0 & 7.0 & 1.3 \\
\hline
512 & C, L, G  &  \checkmark &  &  & 44.9 & 7.3 & 1.5 \\
512 & C, L, G  &  \checkmark & \checkmark &  & 47.5 & 8.2 & \textbf{1.5} \\
512 & C, L, G  & \checkmark &  \checkmark & \checkmark & \textbf{52.9} & \textbf{9.1} & 1.1 \\
\hline
\end{tabular}
}
\end{table}

\begin{table}
\centering
\caption{NetVLAD night query results after ToDayGAN translation; only for a subset of models from Table~\ref{tbl:main_results} to check generalization potential.}
		\label{tbl:netv_results}
\setlength{\tabcolsep}{4pt}
\resizebox{1.02\columnwidth}{!}{
\begin{tabular}{|c|c|c|c|c!{\vrule width 2pt}c|c|c|}
\hline
 &  &  &  &  &  \multicolumn{3}{c|}{Threshold Accuracy (\%)} \\
Resolution & Discriminators & Rel.-loss? & Dual-Eval.? & L/R? & 5m \ang{10} & 0.5m \ang{5} & 0.25m \ang{2} \\
\hline\hline
286 & C, L  &  &  &  & 12.7 & 1.5 & 0.9 \\
286 & C, L, G  &  &  &  & 30.5 & 4.7 & 0.9 \\
\hline
512 & C, L, G  & \checkmark &  \checkmark & \checkmark & 45.6 & 6.4 & 1.3 \\
\hline
\end{tabular}
}
\end{table}

\begin{table}[t]
\centering
\caption{Top overall results for both Night and Night-Rain categories in comparison to other state-of-the-art methods (taken from~\cite{TORSTEN}).}
		\label{tbl:recap_results}
\setlength{\tabcolsep}{4pt}
\resizebox{1.02\columnwidth}{!}{
\begin{tabular}{|c!{\vrule width 2pt}c|c|c!{\vrule width 2pt}c|c|c|}
\multicolumn{1}{c}{} & \multicolumn{3}{c}{\emph{Night}} & \multicolumn{3}{c}{\emph{Night-Rain}} \\
\hline
 &  \multicolumn{3}{c!{\vrule width 2pt}}{Threshold Accuracy (\%)} & \multicolumn{3}{c|}{Threshold Accuracy (\%)} \\
Method & 5m \ang{10} & 0.5m \ang{5} & 0.25m \ang{2}  &  5m \ang{10} & 0.5m \ang{5} & 0.25m \ang{2} \\
\hline\hline
FAB-MAP~\cite{FABMAP} &  0.0 & 0.0 & 0.0  &  0.0 & 0.0 & 0.0 \\
ActiveSearch~\cite{ACTIVESEARCH} &  3.4 & 1.1 & 0.5  &  5.2 & 3.0 & 1.4 \\
CSL~\cite{CSL} &  5.2 & 0.9 & 0.2  &  9.1 & 4.3 & 0.9 \\
NetVLAD~\cite{NETVLAD} &  15.5 & 1.8 & 0.2  &  16.4 & 2.7 & 0.5 \\
DenseVLAD~\cite{DENSEVLAD} &  19.9 & 3.4 & 0.9  &  25.5 & 5.6 & 1.1 \\
\textbf{ToDayGAN (ours)} &  \textbf{52.9} & \textbf{9.1} & \textbf{1.1}  &  \textbf{47.9} & \textbf{12.5} & \textbf{3.2} \\
\hline
\end{tabular}
}
\end{table}

\section{Discussion}

\subsection{Comparing results with baselines}
Our best model using the large training set with left/right images, 512-resolution, three discriminators, relativistic loss, and the flipped-image dual evaluation attains a gain of 2.65x on the 5\textit{m}/\ang{10} threshold over the best DenseVLAD result. The 0.5\textit{m}/\ang{5} category also sees a proportional boost from 3.4\% to 9.1\%. While our 0.25\textit{m}/\ang{2} result also is marginally lower than when not using left/right images, they all seem to be low enough ($\leq$ 1.5\%) that their values are not meaningful.

Testing the best-performing ToDayGAN model directly on the secondary query set of Night-Rain images also works very well, implying our model is robust to the appearance shift. While the relative increases are not as high as the original Night-query's, the absolute accuracies are nearly identical - even higher for the two stricter thresholds.

UNIT's performance is notable in the 5\textit{m}/\ang{10} threshold but poorly for the other two thresholds. UNIT's variational encoding structure tends to result in blurry images, as shown in Figure \ref{fig:res_real}. Lack of low-level detail in the image appears to impair its ability to localize at finer scales, but contains higher-level details sufficient for localizing a general area.

As an additional test, we ran NetVLAD on some of the resulting images as well (see Table \ref{tbl:netv_results}) in order to check the generalization potential of the images for a different comparison method. It turns out to have about the same improvement boost as opposed to directly using NetVLAD (see Table \ref{tbl:recap_results}), suggesting both DenseVLAD and NetVLAD largely rely on the same characteristics - mostly gradients.

We mentioned in Section \ref{rel_oxford} that we attempted to compare our method with \cite{OXFORD}. Due to lack of source code, we could not train their model on our data. Since their method optimizes for SURF features, an approximation of SIFT, it should be suited for DenseVLAD-based localization. The authors did manage to infer our query images on their existing model (trained for a different camera type/view), yet localizing with it performed worse than the naive baseline. So we determined no fair comparison could be made due to the difference in training sets and camera intrinsics.

\subsection{Impact of modifications to ComboGAN}
The first sector of Table \ref{tbl:main_results} ablates the different discriminator combinations to evaluate their contribution to the task. We can see that adding the gradient discriminator just about doubles accuracies, while adding the luminance discriminator improves performance, though to a much lesser degree. Performing the pipeline in RGB rather than just grayscale, which seems to be easier for the networks, regularizes and improves the process invaluably. The use of a combined discriminator is considerably inferior to independent ones. Additionally, note the use of gradient magnitude/orientation as discriminator features unexpectedly behaves much worse in practice compared to just the gradients themselves. There is no obvious explanation, but it points to neural nets having more difficulty dealing with the concepts of gradient angles.
The second sector of Table \ref{tbl:main_results} shows the effects of the non-discriminator factors. The Relativistic-Discriminator loss, Dual-Evaluation procedure, and use of left/right images to increase training set size all improve results.

Lastly, comparing vanilla CycleGAN in Table \ref{tbl:base_results} to the first entry of Table \ref{tbl:main_results}, whose only difference is a multi-scale discriminator scheme, we see an improvement in the stricter thresholds, suggesting the multi-scale architecture handles finer-grained details better, as intended

\section{Conclusion \& Future Work}
In this paper, we have introduced a visual localization system based on image-to-image translations. Our results show that our approach significantly outperforms previous work on the challenging task of localizing nighttime queries against a set of daytime images.

One of the corollaries that can be deduced from our ablation experiments is the partitioning of features for discriminators in a generative-adversarial setup. We find using discriminators tasked with different aspects of a single input image perform better in terms of encouraging the presence of those aspects in generated outputs.

Future work in the field of generative models, in general, can borrow from this very idea. Generated image quality can potentially be improved by using multiple discriminators, each focusing on different image features. Furthermore, these features need not be handcrafted; one could potentially enforce an orthogonality constraint of sorts on the initial features extracted by each discriminator to ensure each concentrates a different aspect of the same input.

\section*{Acknowledgments}
This work was partly supported by ETH Zurich General Fund (OK) and Nvidia through a hardware grant.

\vspace{7pt}
Code for this project is publicly available at \url{https://github.com/AAnoosheh/ToDayGAN}
\vspace{11pt}

\bibliographystyle{plain}
\bibliography{references}

\clearpage
\appendix

\vspace{17pt}

\begin{table}[h]
\centering
\caption{Layer specifications for ComboGAN Generator (Encoder + Decoder) and Discriminator. We use the following abbreviations for brevity: N=Neurons, K=Kernel size, S=Stride size. The transposed convolutional layer is denoted by DCONV. The residual basic block is denoted as RESBLK. (Table taken from \cite{COMBOGAN})}
		\label{tbl:combogan_architecture}
\vspace{3pt}
\scalebox{0.88}{
\begin{tabular}{cl}
\hline
Layer \#                       & Encoders                                \\ \hline
1                              & CONV-(N64,K7,S1), InstanceNorm, PReLU    \\
2                              & CONV-(N128,K3,S2), InstanceNorm, PReLU   \\
3                              & CONV-(N256,K3,S2), InstanceNorm, PReLU   \\
4                              & RESBLK-(N256,K3,S1), InstanceNorm, PReLU \\
5                              & RESBLK-(N256,K3,S1), InstanceNorm, PReLU \\
6                              & RESBLK-(N256,K3,S1), InstanceNorm, PReLU \\
7                              & RESBLK-(N256,K3,S1), InstanceNorm, PReLU \\ \hline
Layer \#					   & Decoders					             \\ \hline
1                              & RESBLK-(N256,K3,S1), InstanceNorm, PReLU \\
2                              & RESBLK-(N256,K3,S1), InstanceNorm, PReLU \\
3                              & RESBLK-(N256,K3,S1), InstanceNorm, PReLU \\
4                              & RESBLK-(N256,K3,S1), InstanceNorm, PReLU \\
5                              & RESBLK-(N256,K3,S1), InstanceNorm, PReLU \\
6                              & DCONV-(N128,K4,S2), InstanceNorm, PReLU  \\
7                              & DCONV-(N64,K4,S2), InstanceNorm, PReLU   \\
8                              & CONV-(N3,K7,S1), Tanh                   \\ \hline
Layer \#					   & Discriminators				             \\ \hline
1 							   & CONV-(N64,K4,S2), PReLU                \\
2 							   & CONV-(N128,K4,S2), InstanceNorm, PReLU \\
3 							   & CONV-(N256,K4,S2), InstanceNorm, PReLU \\
4 							   & CONV-(N512,K4,S2), InstanceNorm, PReLU \\
5 							   & CONV-(N256,K4,S1), InstanceNorm, PReLU \\
6 							   & CONV-(N1,K4,S1)          			     \\ \hline
\end{tabular}
}
\end{table}

\vspace{42pt}

\begin{figure}[ht]
\centering
\includegraphics[width=0.94\linewidth]{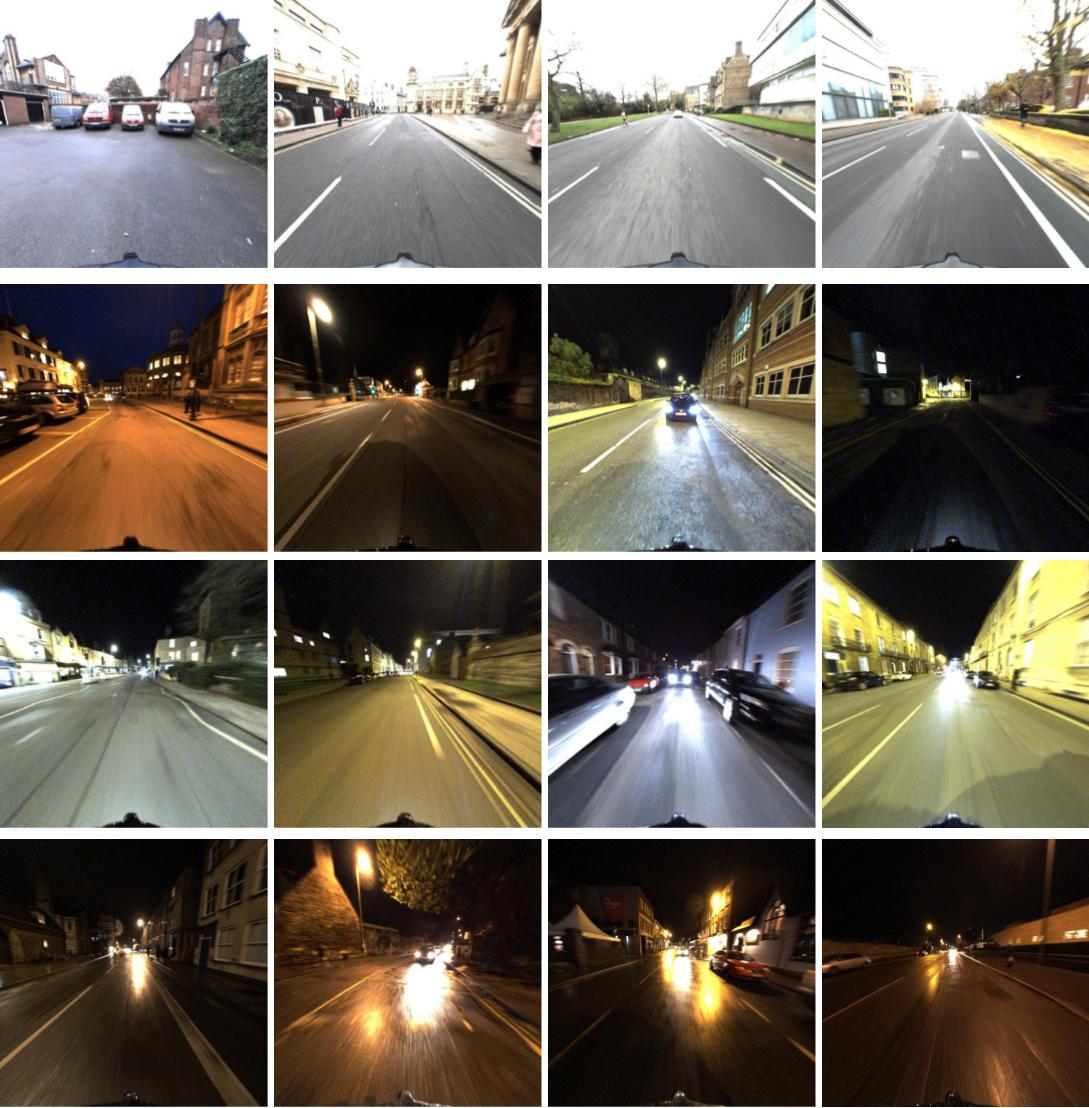}
\caption{Top to bottom row: Sample images from the Daytime set, Night-training set, Night-query set, and Night-Rain query set.}
\label{fig:samples}
\end{figure}

\begin{figure}
\centering
\includegraphics[width=\linewidth]{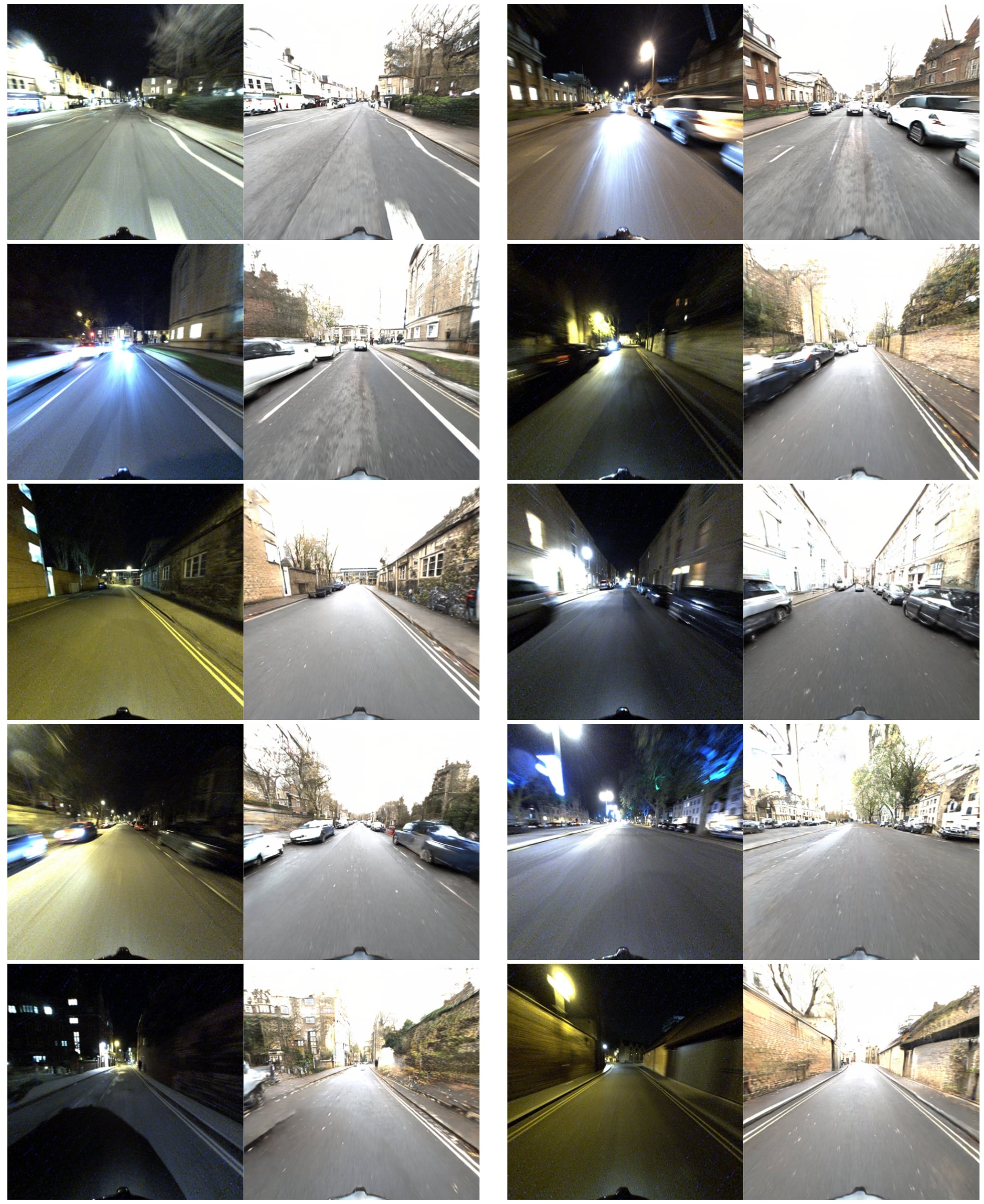}
\caption{Additional visuals for night-to-day translation using ToDayGAN.}
\label{fig:extra_samples}
\end{figure}

\begin{figure}[ht]
\centering
\includegraphics[width=.27\linewidth]{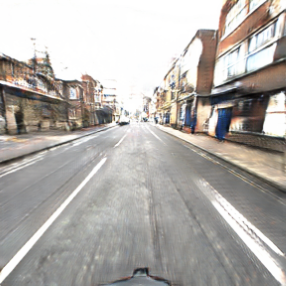}\,
\includegraphics[width=.27\linewidth]{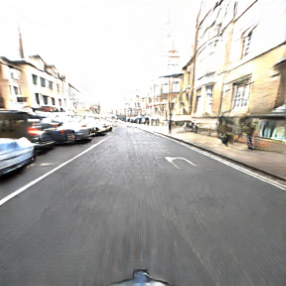}\,
\includegraphics[width=.27\linewidth]{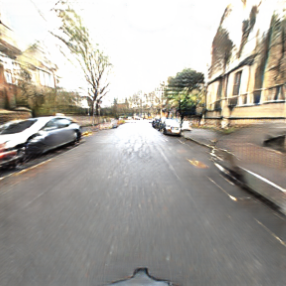}\smallskip

\includegraphics[width=.27\linewidth]{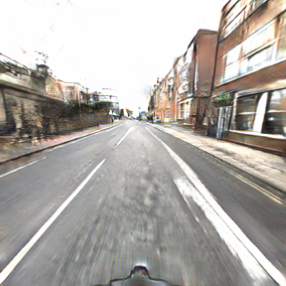}\,
\includegraphics[width=.27\linewidth]{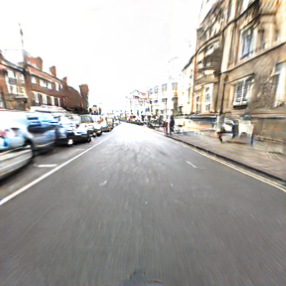}\,
\includegraphics[width=.27\linewidth]{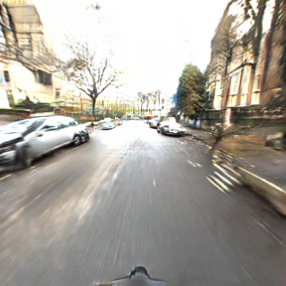}\smallskip

\includegraphics[width=.27\linewidth]{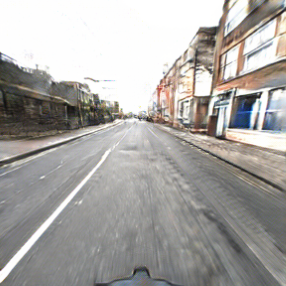}\,
\includegraphics[width=.27\linewidth]{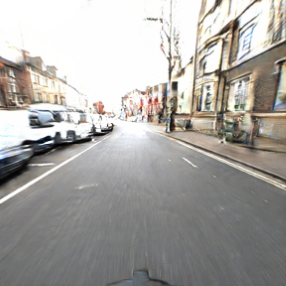}\,
\includegraphics[width=.27\linewidth]{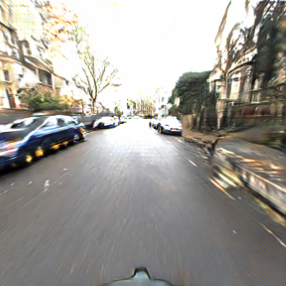}\smallskip

\includegraphics[width=.27\linewidth]{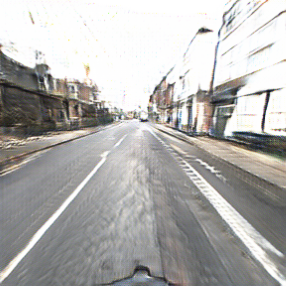}\,
\includegraphics[width=.27\linewidth]{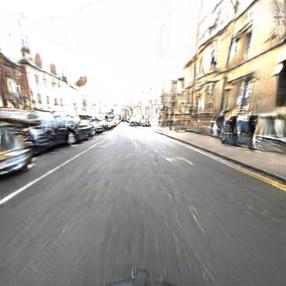}\,
\includegraphics[width=.27\linewidth]{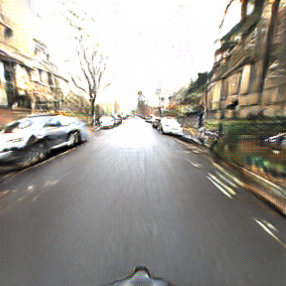}

\caption{Daytime translation of the real night images from Figure \ref{fig:res_real} using different discriminator setups ($286 \times 286$ resolution). Using same notation as in Table \ref{tbl:main_results}, from top to bottom row: C, CL, CLG, and C+L+G.}
\label{fig:res_clg}
\end{figure}

\end{document}